\documentclass[conference]{IEEEtran}
\usepackage{blindtext, graphicx}
\usepackage{hyperref}
\usepackage[style=ieee]{biblatex} 
\ifCLASSINFOpdf
\else
\fi
\begin{document}
%
\title{Neuroevolution for RTS Micro}

\author{\IEEEauthorblockN{Aavaas Gajurel\IEEEauthorrefmark{1},
Sushil J Louis\IEEEauthorrefmark{2},  Daniel J M\'endez\IEEEauthorrefmark{3} and
Siming Liu\IEEEauthorrefmark{4}}
\IEEEauthorblockA{Department of Computer Science and Engineering,
University of Nevada Reno\\
Reno, Nevada\\
Email: \IEEEauthorrefmark{1}avs@nevada.unr.edu,
\IEEEauthorrefmark{2}sushil@cse.unr.edu,
\IEEEauthorrefmark{3}dmendez@nevada.unr.edu,
\IEEEauthorrefmark{4}simingl@unr.edu}}

\maketitle

\begin{abstract}
	This paper uses neuroevolution of augmenting topologies to evolve control tactics for groups of units in real-time strategy games. In such games, players build economies to generate armies composed of multiple types of units with different attack and movement characteristics to combat each other. This paper evolves neural networks to control movement and attack commands, also called micro, for a group of ranged units skirmishing with a group of melee units. Our results show that neuroevolution of augmenting topologies can effectively generate neural networks capable of good micro for our ranged units against a group of hand-coded melee units. The evolved neural networks lead to kiting behavior for the ranged units which is a common tactic used by professional players in ranged versus melee skirmishes in popular real-time strategy games like Starcraft. The evolved neural networks also generalized well to other starting positions and numbers of units. We believe these results indicate the potential of neuroevolution for generating effective micro in real-time strategy games.
\end{abstract}


\section{Introduction}
Real Time Strategy (RTS) games are a genre of multi-player video games where players take actions concurrently and the underlying game world dynamically changes over time. The overarching objective of the the game is to establish a position capable of defending against and destroying opponents. Actions in the game can be largely divided into two modes: "macro" and "micro". Macro management relates to long term  strategic decisions and is concerned with resource gathering, spending those resources on research, deciding on the type and number of units to build, and in building those units. Micro management is concerned with quick and short term tactical control of units usually during a skirmish between a group of friendly units against an opponent’s group. RTS game environment are a partially observable and imperfect information environment due to a restricted view through the camera on a part of the whole map and a "fog of war" which hides information form parts that have not been explored. Players have to control numbers of units ranging from tens to hundreds while simultaneously moving the camera around, deciding which units or unit factories to build, selecting units, scouting, and exploring. The state space of typical RTS game like Starcraft is estimated to be more than 
$10^{50^{36000}}$ using a conservative branching factor of $10^{50}$ for each frame in a $25$ minute game~\cite{ontanon2013survey}. Consequently, RTS games provide a challenging platform  for  testing  machine  learning  approaches~\cite{buro2004call}.

This paper focuses on generating artificial agents capable of good micro control in RTS games. Micro requires quick decision making and fast successive actions to control both movement and attack commands for units in a group. There are multiple types of units, each with its own advantages and disadvantages. Each unit has unique, well-defined characteristics regarding its capabilities, like weaponry, range, speed, maneuverability and others. Good micro can be a deciding factor in a skirmish between two groups with similar characteristics and the player has to consider the attributes of both friendly and enemy units to choose an effective tactic for the particular scenario. The complexity of the different ways in which any unit group can be controlled is as a result challenging, particularly since directives have to be provided in quick reactive time-frames. 

RTS games have been used as an environment for AI research and various approaches towards automation of different aspects of RTS game playing have been explored~~\cite{ontanon2013survey}. Approaches like reinforcement learning, scripting, and search, among others, have been used with the end goal of creating a fully automated, human-comparable RTS player~~\cite{robertson2014review}. Previous work has explored using Genetic Algorithms (GA) to search for an optimal combination of parameters, which are then used in Potential Fields (PF) and Influence Maps (IM) equations to control the tactical actions of skirmishing units~~\cite{simingpaper}. Our research builds on this previous work in RTS game AI, but takes a different approach. Rather then having a set of parameterized control algorithms, or potential fields, for controlling movement, we explore the feasibility of evolving a neural network to perform good micro. In particular, we explore using Neuro-Evolution of Augmented Topologies (NEAT)~~\cite{neatpaper} to evolve neural networks to effectively and autonomously control units for skirmishes in RTS games.

NEAT evolves both the structure and connection weights of a neural network by utilizing genetic algorithm principles~\cite{holland1992adaptation} and applying them to a population whose chromosomes represent different instances of networks being explored. Thus, the NEAT approach encodes both the structure and weights of a neural network that tries solve a problem. NEAT incorporates historical markings, speciation and starts complexification from a minimal network. There is good empirical evidence that NEAT can evolve robust solutions for non-trivial problems~~\cite{hausknecht2014neuroevolution}.

We feed the evolving network with the relative positions of all units in the arena along with the unit's internal state as inputs. There are three outputs. Two specify a relative 2D position $(\delta x, \delta y)$ to move towards and third represents a binary value that determines if we move or attack. 

Preliminary results, on an underlying RTS-physics implementing simulation, show that NEAT can evolve networks for micro control of ranged units against a group of melee units. The evolved network generated kiting behaviour for ranged units (copied from vultures in Starcraft) which allowed five vultures to eliminate twenty-five zealots (a strong melee unit) without suffering any damage. We evolved our network on ten different starting configurations differing in the initial positions and numbers of zealots, and tested the evolved network on configurations with varying numbers of zealots from one to thirty. Our results indicate that evolved networks generalize well to different starting configuration and varying numbers of vultures. We then moved to the recently released Starcraft II (SCII) game API and were able to show that NEAT can evolve good micro on a simple, flat Starcraft II map with no obstacles. Although the networks volved in SCII are not as effective as those that evolve in our simulation, when pitting 5 hellions  against 25 zealots, hellions learn to kite and can on average destroy close to a majority of zealots. 
\footnote{Like vultures in Starcraft, hellions are also relatively fragile, longer ranged, and fast Starcraft II units.}
As before, the NEAT networks generalized to different numbers of zealots and to different starting locations. We believe our method of network representation can be extended to incorporate new inputs to be applicable to more complex micro scenarios. In addition, we believe that we can significantly improve NEAT evolved micro in SCII with more computational resources.

The remainder of this paper is organized as follows. Section ~\ref{RelatedWork} describes previous approaches related to our current work, section ~\ref{methodology} describes the neural network representation, evolution configurations and NEAT setup. In Section ~\ref{results}, we describe our generalization results and lastly in section ~\ref{conclusion} we draw our conclusions and explain possible future approaches.

\section{Related Work} {\label{RelatedWork}}

Significant work has been done over the years in the field of designing effective RTS AI players using different techniques ~~\cite{ontanon2013survey}. Buckland et al. described a rule based approach in his book~\cite{buckland2002ai} and Rabin and Steve ~\cite{rabin2014ai} explained scripted agents which is a general approach used by bots that play in starcraft AI ladder matches. Weber and Mateas ~\cite{weber2009data} explored using data mining on gameplay logs to predict the strategy of an opponent. A tree based search approach was used by Churchill, Saffidine and Buro who utilized transition tables to generate trees of actions and performed alpha beta pruning to create agents for 8 vs 8 unit skirmish ~\cite{churchill2012fast}. Others have also tried reinforcement Learning: Wender and Watson ~\cite{wender2012applying} used Q learning and Sarsa to develop a fight or retreat agent. Shantia, Begue and Wiering ~\cite{shantia2011connectionist} applied reinforcement learning on neural networks by using neural-fitted and online versions of the Sarsa Algorithm where they implemented a state space representation similar to ~\cite{wender2012applying}. Vinyals et al. ~\cite{vinyals2017starcraft} applied deep reinforcement learning in a Starcraft II environment to train neural networks using gameplay data from expert players. Their representation used raw image features corresponding to the game display called feature maps and they provide baseline results for convolutional, longterm shortTerm memory and random policy based agents.

Potential Fields (PFs), which has been widely used for robot navigation and obstacle avoidance [21][22]~\cite{khatib86} have also been used for micro. Hagelback and Johansson ~\cite{hagelback} presented a multi-agent potential field based bot architecture for the RTS game ORTS ~\cite{buro} which incorporated PFs into their unit AI. More recent work has focused on combining PFs with Influence Maps (IM) to represent unit and terrain information. In this context, an influence map is a grid superimposed on the virtual world where each cell is assigned a value by an IM function, which is used by an AI to determine desired actions ~\cite{uriarte2012kiting} ~\cite{simingpaper}. Coevolution was used by Avery and Louis in ~\cite{avery2010coevolving} to develop micro behaviours by coevolving influence maps for team tactics and in ~\cite{miles2006co} where they cooevolved influence map trees(IMT) and show that evolved IMTs displayed similar behaviours to hand coded strategies.

NEAT has been applied to dynamic control tasks like double pole balancing without velocity information ~\cite{neatpaper} where it could evolve a robust control policy. It has also been applied to evolving walking gaits for virtual creatures ~\cite{allen2009complex} and steering control for driving agents ~\cite{cardamone2009evolving} ~\cite{duran2005driving}. NEAT has also been applied to evolve video game playing agents for games like Ms. Pac-Man~\cite{schrum2014evolving} and Tetris~\cite{gillespie2017comparing} and has been shown to be applicable to general Atari gameplaying ~\cite{hausknecht2014neuroevolution}. Board games like 2048 ~\cite{boris2016evolving} and Go ~\cite{stanley2004evolving} have also been shown to be within reach.

NEAT and its realtime variant rtNEAT have been used to tackle different aspects of RTS games. Olesen et al. ~\cite{olesen2008real} used NEAT and rtNEAT to control the macro aspects of the game to match the difficulty of the opponent. Gabriel et al. ~\cite{gabriel2014neuroevolution} used rtNEAT to evolve a multi-agent system for Brood war agents based on ontological templates, where they show that their hierarchical method could be used to evolve good micromanagement tactics. NEAT for RTS micro control was applied in ~\cite{zhen2013neuroevolution} where the authors approached the problem by having a neural network control a combat unit's fight or flee decision, based on various entity properties like weapon cooldown, remaining health, weapon range, enemy weapon range, number of allies in range and number of enemies in range. They used NEAT and rtNEAT and had the fight or flee logic hard-coded for the network to activate; which  differs from the approach in this paper where we are directly trying to control unit movement based on the position of friendly and enemy units around the entity being controlled, without further structure.

\section{Methodology} \label{methodology}

There are different aspects of micro game-play that can be controlled for an entity, such as movement, whether to attack, when to flee, and other such unit specific actions. Controlling all aspects of a micro engagement is therefore a complex endeavor. In this research, we focus on entity movement and firing control. Movement can be further classified into long and short range, based on the timescale within which the action must execute. Longer routes pertain to long distance movement of units, say from a player's base to an enemy's base, while short duration movement, like kiting, are finer tactical movements done in shorter periods of time. Kiting is a strategy that is used by speedy ranged units against slower melee units, where the ranged units fire, run out of range, turn back, fire, and run back out of range again and again avoiding damage to themselves while damaging the enemy. We are trying to evolve networks which can perform similar tactics for ranged units against melee units. 

We next describe the NEAT evolutionary algorithm, the neural network representation for NEAT, and the experimental setup used for evolution.

\subsection{Neuro-Evolution of Augmenting Topologies}

NEAT is a robust algorithm for evolving neural networks based on genetic algorithm principles. NEAT attributes it's robustness to three aspects, specifically that it starts complexifing from minimal structure, that it leverages speciation and its use of historical markings in the genome for crossover and speciation~\cite{neatpaper}. NEAT allows for continuous complexification by allowing mating together with mutation to fully modify the resulting network and a number of different kinds of mutation are used~\cite{neatpaper}.

Neat evolves a neural network from inputs and outputs specified by the problem domain. We specify the inputs and outputs in more detail below. 

\begin{figure}[h]
\centering
\includegraphics[width=0.4\textwidth]{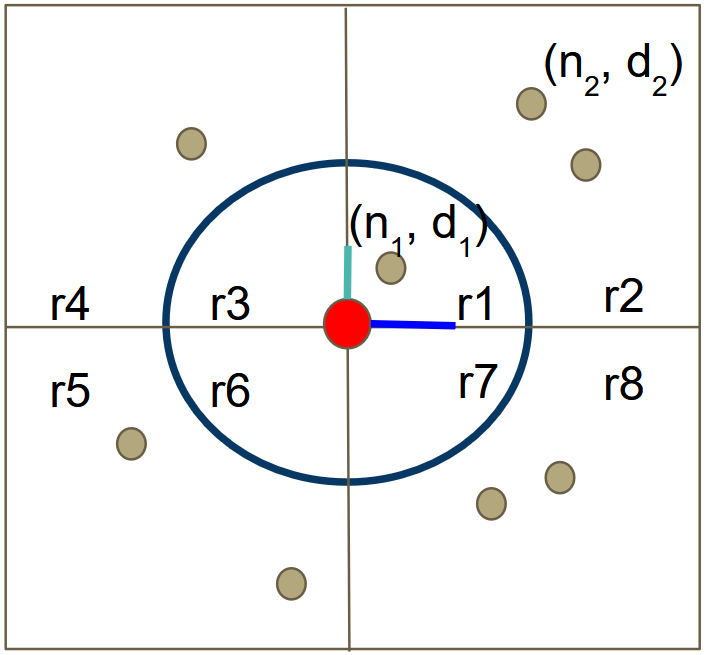}
\caption{Input and Output Representation}
\label{fig:quadrant}
\end{figure}

Our neural network inputs can be divided into two classes according to the type of information they represent. The first class deals with spatial information and describes the relative position of all entities on the map. In order to be able to represent units consistently and uniformly, regardless of the number, we followed an approach whereby we divide the visible world into regions relative to the current position of the unit being controlled. Figure~\ref{fig:quadrant} shows the spatial information being fed into the neural network.
In our representation, the world around the entity is divided into 4 quadrants with the entity at its center. The four quadrants are further divided into eight regions separated by the attack range of the unit as shown by the labels R1 to R8 in Figure~\ref{fig:quadrant}. Each regions then corresponds to four inputs in the network:
\begin{enumerate}
    \item the number of enemy units
    \item average distance of enemy units
    \item number of friendly units and 
    \item average distance of friendly units in the region
\end{enumerate}

Next, we have four inputs indicating map boundaries. These inputs provide distance from the entity to north, south, east and west boundaries correspondingly. The second class of inputs, feeds the entity with some of the entity's own internal state. The internal features that we have considered are

\begin{enumerate}
    \item the current health
    \item weapon cool-down
    \item current fire or move state and 
    \item a recurrent input which is the previous attack/move output from the network
\end{enumerate}

Thus, in total the neural network that controls the movement of friendly units in our environment has 40 inputs, 16 that are used to represent the position of all friendly units, 16 that are used to represent, the position of enemy units, 4 boundary sensors and 4 inputs for the internal state as shown in table \ref{tab:inputs_net}. The representation is constructed such that it can capture essential information from different map configurations and number of entities, without having to vary the number of inputs in the network. Once computed, all inputs are scaled between 0 to 1 by representing each value as percentage of a maximum possible value for that input. 
\begin{table}[h!]
\centering
\caption{Neural Network Inputs}
\label{tab:inputs_net}
 \begin{tabular}{||c |l ||} 
 \hline
 id & Type \\ [0.5ex] 
 \hline\hline
 1-8 & enemy avg position per region \\ 
\hline
  9-16 & friendly avg position per region \\ 
 \hline
  17-24 & enemy units per region \\
 \hline
  25-32 & friendly units per region \\
 \hline
  33-36 & boundary sensors for 4 directions \\ 
 \hline
 37 & self cooldown \\ 
 \hline
 38 & self hitpoint \\ 
 \hline
 39 & current attack/move state \\ 
 \hline
 40 & previous attack/move state \\ 
 \hline
\end{tabular}
\end{table}


\subsubsection{Output representation}

The output is represented by two scalars representing a desired $\delta x$ and $\delta y$ coordinate displacement, and one boolean for whether a unit should fire or move at that instant. $\delta x$ and $\delta y$ displacement output are scalar values from 0 to 1 from which we subtract 0.5 and then scale them to go the coordinate position relative to the unit's current position. This allows the output to represent any coordinate around the entity in the region corresponding to the scaling factor. The outputs are then fed into the movement mechanism of the simulation or SCII in order to generate movement. 
The third output is a move or attack command which is a Boolean signal. If the output is greater than 0.5, the entity has to focus on attack and if the output is lower, the entity stops attacking and begins moving to the position signalled by $\delta x$ and $\delta y$ displacement output.

\subsection{Experimental setup}

 Although our physics-simulation used for preliminary results and for experimentation to explore input and output representations runs fast, the simulated physics and entity properties cannot be easily made identical to SCII. This means that micro evolved in our simulation may not transfer well to SCII. In addition, there are differences in the properties of vultures in our simulation, vultures in Starcraft Brood Wars, and hellions in SCII. However, our simulation runs much faster and we can experimentally try multiple representations, input configurations, and NEAT parameters far more quickly than when using the SCII API. We can then start long evolutionary runs within the SCII environment with more confidence.
 
 For our experiments, we choose the zealot as a representative melee unit and either the vulture or hellion as a representative ranged unit. More specifically, for our simulation environment, we copied zealot and vulture properties from the Starcraft BW API~\cite{vulture_param}. When running in SCII, zealots and hellions use SCII properties.  Vultures/hellions and zealots deal comparable damage in each attack but have different attack range and movement speeds. Table~\ref{tab:properties_entity} shows the properties of the units considered in this paper. hellions and vultures, when micro'd well can be strong against zealots because of their greater speed and attack range, which makes it possible for a small number of vultures/hellions to kite a bigger group of zealots to death. We expect our approach to evolve good tactical control that can exploit this strength of vultures/hellions against zealots.

\begin{table}[h!]
\centering
\caption{Properties of zealots and hellions}
\label{tab:properties_entity}
 \begin{tabular}{||l |r r r||} 
 \hline
 Property & Vulture & Hellion & Zealot \\ [0.5ex] 
 \hline\hline
 Hit-points & 80 & 90 & 100 \\ 
 \hline
  Damage & 20 & 13 & 16 \\ 
 \hline
  Attack Range & 5 & 5 & 0.1 \\ 
 \hline
  Speed & 4.96 & 5.95 & 3.15 \\ 
 \hline
  Cool-down & 1.26 & 1.78 & 0.857 \\ 
 \hline
\end{tabular}
\end{table}

NEAT evolves the network across generations based on the fitness of the network. To evaluate the fitness of the neural network, we used two different environments: the Starcraft II game and our own simulation of the Starcraft environment which is tailored to capture the micro combat aspects of Starcraft and can be run without graphics for significant speedup. 

\begin{figure}[h]
\centering
\includegraphics[width=0.45\textwidth]{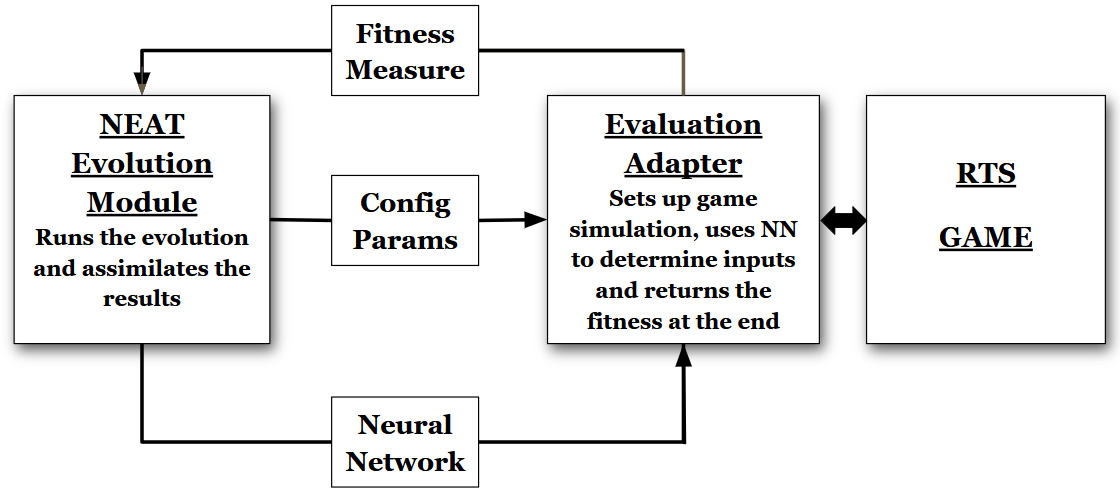}
\caption{Architecture Diagram}
\label{fig:arch}
\end{figure}

Our experimental setup had three main components, the NEAT evolution module, the Simulation adapter and the game itself, which could either be Startcraft II or our own simulation of the game. Neat is concerned with running the evolution by assimilating the fitness results and generating networks. We used the SharpNeat implementation of NEAT by Colin Green~\cite{sharpneat} for the evolution module and adapted it for our purpose. A simulation adapter is the mediator between the evolutionary algorithm and the game which we implemented using sockets to be able to run the game simulations in parallel. It gets the configuration from the NEAT module and sets up the game, it also gets a neural network configuration from the evolution module and feeds inputs with the current game state into the network and uses the output from the network to feed the game and move entities. The adapter returns the final fitness after running the simulation which ends when one of the players has no units left or after a set number of frames. The game component can be either Starcraft or our combat simulation. The architecture diagram of the components is shown in figure \ref{fig:arch}.

\subsubsection{Map configurations}

We choose five different unit spawn configurations that determine entity starting positions. These are diagonal, side by side, surround, surrounded and random. In diagonal, opposing sides spawn in groups diagonally on a square map. Similarly, in the side by side configuration, entity's appear along the same $y$-coordinate separated by a distance. In surrounded, vultures or hellions appear at the center in a group surrounded by number of zealots and the opposite is true for surround - hellions or vultures surround zealots. We also experimented on random spawning locations for all units of both players. We kept the number of vultures or hellions constant at five ($5$) but randomly varied the number for zealots for different configurations. In the rest of the paper, we mean vultures or hellions when we use the term ranged units. 

\subsubsection{Fitness Function}

We used a fitness function that considers both the damage received and the damage dealt by the our evolving ranged units (vultures and hellions). Over evolutionary time, fitness gradually grows as the ranged units get better at damaging zealots and at evading attacks. At the end of each game run, we sum the remaining hitpoints for both zealots $(Hz)$ and ranged units $(Hh)$ and subtract the remaining hitpoints of zealots from the remaining hitpoints of the ranged units. We add the maximum hitpoints for all zealot units so that the fitness function is always positive.

For number of starting zealots $Nz$, remaining zealots $Rz$, remaining hellions $Rh$, and maximum hitpoint of zealot $Hzmax$, fitness $F$ is calculated as:

$$F = \sum_{n=1}^{Nz} Hzmax_n + \sum_{n=1}^{Rh} Hh_n -  \sum_{n=1}^{Rz} Hz_n $$

We should note that as the hit points of both zealots and ranged units are similar, with increasing numbers of zealots and low numbers of ranged units, this fitness function leans towards giving more weight to damage done than damage received. This could be better balanced in various ways. For example, by multiplying the sum of hitpoints of hellions by a balancing factor. Nevertheless, we found that the current configuration performed well during our experimentation phase.

\section{Results and discussion} \label{results}

We experimented with two different RTS game environments. The Starcraft II game, and our simulation of the RTS environment particularly developed for fast simulation of skirmishes. Below, we describe our experiments and  analyze the results for each.  

\subsection{Simulation Results}

\begin{table}[h!]
\centering
\caption{Hyper-parameters For NEAT Evolution }
\label{tab:par_simulation}
 \begin{tabular}{||l |r| r||} 
 \hline
 Property & Simulation & Starcraft II \\ [0.5ex] 
 \hline\hline
 Population & 50 & 50\\ 
 \hline
  Generations & 100  & 100\\ 
 \hline
  Species & 5 & 5 \\ 
 \hline
  Initial Conn Probability  & 0.2 & 0.1 \\ 
 \hline
  Elitism Proportion  & 0.2 & 0.2\\ 
 \hline
  Selection Proportion  & 0.2 & 0.2\\ 
 \hline
 Asexual Offspring Proportion  & 0.5 & 0.8 \\ 
 \hline
 Sexual Offspring Proportion  & 0.5 & 0.2 \\ 
 \hline
  Inter-species  Mating  & 0.01 & 0.01\\ 
 \hline
  Connection Weight Range  & 5 & 7 \\ 
 \hline
  Probability Weight Mutation  & 0.95  & 0.95\\ 
 \hline
  Probability Add Node & 0.01 & 0.02\\ 
 \hline
  Probability Add Connection   & 0.025 & 0.04\\ 
 \hline
  Probability Delete Connection   & 0.025 & 0.025\\ 
 \hline
\end{tabular}
\end{table}

In our first set of experiments using our simulation environment, we evolved vultures against a larger group of zealots. Zealots in our simulation, use a hand coded AI which controls each unit as follows: pursue the nearest vulture and attack when it is in range. Both zealots and vultures were given complete map vision - there was no fog of war and thus they did not have to explore the map and could start pursuing their enemy right away. Note that this is a significant difference from SCII.

We ran NEAT on a population size of 50 individuals for 100 generations. The following results are average of 10 different runs of a complete evolutionary epoch, started with different random seeds. Various hyper-parameters that we used for the evolution are noted in table \ref{tab:par_simulation}.  Each genome was evaluated based on a complete run of a test configuration, which consisted of 10 different spawning locations with different number of zealots and vultures. We sum the fitness for each of the 10 different training configurations to get the final fitness, which is then forwarded to the NEAT module. We run each scenario until one of the player looses all his units or for a maximum number of frames. We had the option to run our simulation without the graphics rendering which significantly decreased running time compared to SCII.

Initially, we tried to evolve agents only based on a single test configuration of the map, but results showed that they did not generalize well to new scenarios. Using the sum of fitnesses from different configurations led to good generalization across different map configurations and different numbers of units. The 10 different test cases are a sample from the the possible configuration space of different number of zealots and 5 different starting configuration. The training scenarios are listed below.

\begin{enumerate}
    \item Diagonal, 25 zealots
    \item Reversed Diagonal, 20 zealots
    \item Side by Side, 10 zealots
    \item Reversed Side by Side, 15 zealots
    \item Surround, 20 zealots
    \item Surround, 10 zealots
    \item Surrounded, 20 zealots
    \item Surrounded, 25 zealots
    \item Random, 15 zealots
    \item Random, 25 zealots
\end{enumerate}

In our simulation environment, the average number of generations over ten runs, needed to find the best individual was $80$ and the average best fitness was $96\%$ of the maximum fitness possible. We found that the the evolved vultures learned kiting or to hit and run, against the group of zealots. Kiting is an effective tactic for speedy ranged units against slow melee units as explained earlier.

\begin{figure}[h]
\centering
\includegraphics[width=0.45\textwidth]{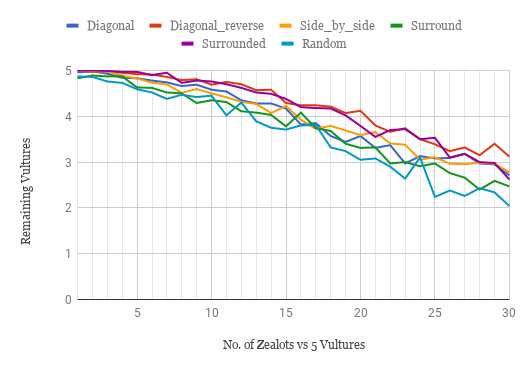}
\caption{Remaining Vultures corresponding to increasing zealot numbers}
\label{fig:mysim_rem_vult}
\end{figure}
After evolving neural networks to control vultures with kiting ability against groups of zealots, we tested for the generalizability of our result against scenarios that the Vultures did not encounter during the training phase. For each possible starting configuration, we varied the number of starting zealots from 1 to 30 while the number of Vultures was always constant at 5. Here, we note that Vultures were only evolved against the group of 10, 15, 20 and 25 zealots thus, their performance against different number of zealots shows the robustness of the evolved network.

\begin{figure}[h]
\centering
\includegraphics[width=0.45\textwidth]{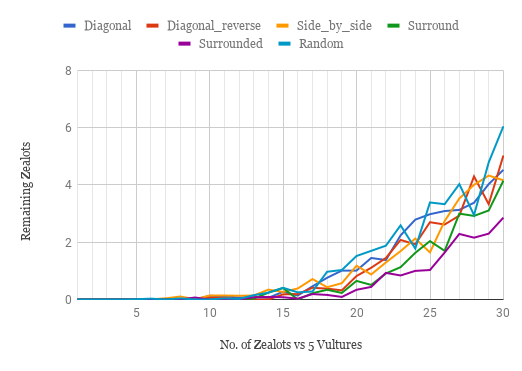}
\caption{Remaining zealots corresponding to increasing zealot numbers}
\label{fig:mysim_rem_zeal}
\end{figure}

Results of our generalizability tests are shown in Figures~\ref{fig:mysim_rem_vult} and~\ref{fig:mysim_rem_zeal}. The vertical axis represents the number of units remaining at the end of each game run and the horizontal axis represents the number of zealots against which the five vultures skirmish. We present the results for 6 different starting positions, each starting position averaged over ten runs. Figure~\ref{fig:mysim_rem_vult} shows the number of vultures remaining when times runs out while Figure~\ref{fig:mysim_rem_zeal} shows the number of zealots remaining. As shown in Figure~\ref{fig:mysim_rem_vult}, we see that vultures' performance smoothly decreases as the number of zealots increases. The number of vultures never goes below two, a good indication of the robustness of evolved networks.  

Generalization with respect to damage done is shown in figure \ref{fig:mysim_rem_zeal} where we note that the zealots are completely destroyed by vultures till the number of starting zealots rises above 13. The number of surviving zealots then gradually increases across all our scenarios. The number of remaining zealots never rises above six another good indicator of micro quality and robustness. 

We also found that we must provide a number of different training configurations in order to evolve robustness. Evolving with only one configuration, results in much less robust networks whose performance might jump form high to low or low to high when changing the number of zealots even by as little as one zealot. In one case removing a single zealot significantly {\bf decreased} vulture performance. This result is not unexpected since much research in neural networks and other machine learning has shown similar effects.


\subsection{Starcraft II Results}

In Starcraft II, we evolved hellions which are a ranged units that can do splash damage against zealots which are strong melee units. 
We control the movement and attack commands for the hellions after translating the outputs from NEAT evolved neural networks to the commands for Starcraft II using the API. We ran the game at the top speed of 16. As Starcraft II runs relatively slow even at top speed, we ran on 15 machines for 24 hours and would have preferred to have a much larger cluster.

For StarcraftII experiments, we ran on a population size of 50 for 100 generations and the results averaged over 10 times for the final results. We use the same NEAT parameters as for our simulation and given in~\ref{tab:par_simulation}. Unlike our simulation, we used the sum of fitness for only three different configurations to get the individual fitness. The three configurations and corresponding number of zealots are listed below; the number of hellions is always five.

\begin{enumerate}
    \item Diagonal, 25 zealots
    \item Random, 20 zealots
    \item Side by Side, 15 zealots
\end{enumerate}

Over 10 runs, the average number of generations needed to find the best individual was $85$ and the average best fitness was $88\%$ of the maximum fitness possible. The evolved network  also displayed kiting behaviour against the zealots - similar to our findings from the simulation approach.

\begin{figure}[h]
\centering
\includegraphics[width=0.45\textwidth]{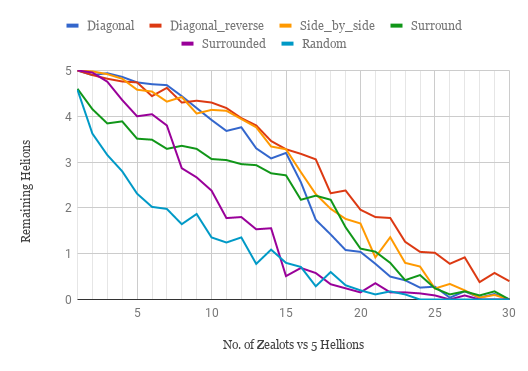}
\caption{Remaining hellions corresponding to increasing zealot numbers}
\label{fig:sc2_rem_helli}
\end{figure}

\begin{figure}[h]
\centering
\includegraphics[width=0.45\textwidth]{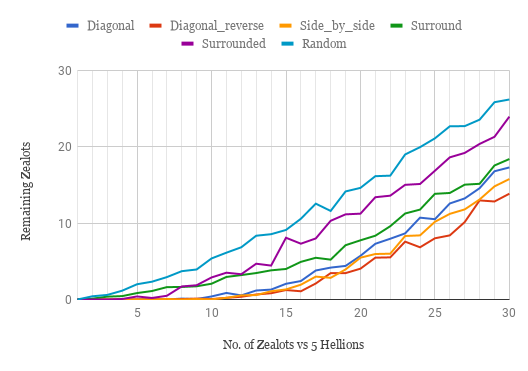}
\caption{Remaining zealots corresponding to increasing zealot numbers}
\label{fig:sc2_rem_zeal}
\end{figure}

We tested for the generalizability of the evolved networks in similar fashion to the tests for then simulation environment. That is, we tested the best evolved network against new configurations and with varying number of zealots. Here, we note that hellions only evolved against groups of 15, 20 and 25 zealots, and on only three configurations. Generalization results are shown in Figure~\ref{fig:sc2_rem_helli} and \ref{fig:sc2_rem_zeal}. The vertical axis represents the number of units remaining at the end of each game run and the horizontal axis represents the number of zealots remaining when skirmishing with five hellions. We present the results for six different starting positions with the number of zealots varying from 1 to 30. We ran the simulation for five runs on the same starting configuration to get the average number of remaining units. We expect to do more runs as computational resources allow.

As shown in figure \ref{fig:sc2_rem_helli}, we see a downward trend for then number of remaining hellions starting from 5. However, unlike in our simulation, the trends are different for different starting configurations.  Diagonal and side by side show good performance across different numbers of zealots while random and surrounded perform comparatively lower. The gradual decrease is expected as the hellions get overwhelmed by the increasing number of zealots. Still, we can see from the graph that hellions are generalizing well with respect to damage received against different number of zealots and different starting positions. 

Generalization with respect to damage done is shown in figure \ref{fig:sc2_rem_zeal} where we again note that the hellions perform well for diagonal and side by side scenarios while performing comparatively lower in random and surrounded scenarios. The number of remaining zealots for each scenario only gradually increases on these previously unseen scenarios and indicates generalizability of our evolved result. 

\subsection{Comparison of Results}

We have shown that the NEAT generated neural networks from from Starcraft II and our simulation were able to generalize with respect to different starting positions and different numbers of zealots. Ranged units performed well against smaller numbers of zealots and performance decreased with increase in number of zealots for both environments. The gradual progression of values for different series without major deviance indicates that the evolved network is robust against changes in both starting position and number of zealots. Videos at \underline{\url{https://www.cse.unr.edu/~aavaas/Micro.html}} serve well to indicate the quality and robustness of the evolved micro. 

Neat was able to evolve kiting behaviour in both our simulation and in the Starcraft II environments but there are some differences between the results from two environments. The evolved network in Starcraft II seemed to perform comparatively worse than our simulation. We believe the fewer training configurations, the increased complexity of SCII, and differences in the AI we evolved against, account for these differences. 

\section{Conclusion and Future Work} \label{conclusion}

Our research focused on using NEAT to evolve neural networks that could provide robust control of a squad in an RTS game skirmish. We showed that our representation of the game state provided to NEAT sufficed to evolve high performing micro, while training on a variety of scenarios leads to more robust and high performing micro. 
The evolved networks generalized well to different numbers of opposing units and different starting configurations. 

We used our own simulation environment for initial experimentation and exploration. Because our simulation runs much faster than Starcraft II, we were able to explore multiple representations and evolutionary parameters to hone in good representations and parameters. We then used this experience to try reproduce our results in the popular RTS game. Starcraft II. Here, we ranged hellions evolved kiting behaviour against melee zealots - like in our simulation environment and meeting our expectations. 
We believe these results show that NEAT holds promise as a potential approach to evolving RTS game micro. 

With a general neural network representation and with NEAT, we think that our approach can be effectively extended to approach more complex scenarios and group configurations. We are working on probabilistic activation model for outputs: we can consider the output of the neural network as the probability of it being active rather than comparing it against the threshold. Using recurrent neural networks would enable incorporating sequential strategies spanning multiple time frames. In addition, we will be adding more game state information about opponents, considering a multi-objective formulation of the fitness function, and obtaining and using much larger computational resources. 


%




\ifCLASSOPTIONcaptionsoff
  \newpage
\fi



\printbibliography 

%

\begin{IEEEbiography}[{\includegraphics[width=1in,height=1.25in,clip,keepaspectratio]{picture}}]{John Doe}
\blindtext
\end{IEEEbiography}




\end{document}